\pdfoutput=1

\documentclass[11pt]{article}

\usepackage[preprint]{coling}

\usepackage{times}
\usepackage{latexsym}
\usepackage{amsmath} 
\usepackage{xcolor}

\usepackage[T1]{fontenc}

\usepackage[utf8]{inputenc}

\usepackage{microtype}

\usepackage{inconsolata}
\usepackage{xspace}
\usepackage{ulem}

\usepackage{graphicx}

\newcommand{\gptfour}{\texttt{PE-GPT4}\xspace}
\newcommand{\gptthree}{\texttt{PE-GPT3.5}\xspace}
\newcommand{\llama}{\texttt{PE-Llama2}\xspace}
\newcommand{\mix}{\texttt{PE-Mixtral}\xspace}
\newcommand{\uni}{\texttt{UniEval}\xspace}
\newcommand{\dialogrpt}{\texttt{DialogRPT}\xspace}
\newcommand{\prompteval}{\texttt{PromptEval}\xspace}

\definecolor{mygreen}{RGB}{0,128,0}
\definecolor{myred}{RGB}{200, 0, 0}
\definecolor{myyellow}{RGB}{200, 128, 0}

\title{Measuring the Robustness of Reference-Free Dialogue Evaluation Systems}

 \author{ Justin Vasselli \and Adam Nohejl \and Taro Watanabe \\
         Nara Institute of Science and Technology \\
         \texttt{\{vasselli.justin\_ray.vk4, nohejl.adam.mt3, taro\}@is.naist.jp}}

\begin{document}
\maketitle
\begin{abstract}
Advancements in dialogue systems powered by large language models (LLMs) have outpaced the development of reliable evaluation metrics, particularly for diverse and creative responses. We present a benchmark for evaluating the robustness of reference-free dialogue metrics against four categories of adversarial attacks: speaker tag prefixes, static responses, ungrammatical responses, and repeated conversational context. We analyze metrics such as \dialogrpt, \uni, and \texttt{PromptEval}—a prompt-based method leveraging LLMs—across grounded and ungrounded datasets. By examining both their correlation with human judgment and susceptibility to adversarial attacks, we find that these two axes are not always aligned; metrics that appear to be equivalent when judged by traditional benchmarks may, in fact, vary in their scores of adversarial responses. These findings motivate the development of nuanced evaluation frameworks to address real-world dialogue challenges.~\footnote{\url{https://github.com/JVasselli/dialogue-metric-robustness}}
\end{abstract}

\section{Introduction}

Despite significant advancements in dialogue systems driven by large language models (LLMs), the development of reliable evaluation metrics remains an open challenge, particularly in capturing the nuances of diverse and creative responses. The wide range of valid responses to a given dialogue context often renders reference-based metrics ineffective, as they rely on a limited set of references which fail to capture the full spectrum of acceptable responses \citep{liu-etal-2016-evaluate}. This shortcoming risks unfairly penalizing creative or contextually appropriate responses.

Reference-free metrics such as \dialogrpt~\citep{gao-etal-2020-dialogue} and \uni~\citep{zhong-etal-2022-towards}, offer a promising alternative \citep{ren-etal-2023-c}. However, their reliability has not been fully tested. For instance, \citet{hicke-etal-2023-assessing} demonstrated that \dialogrpt is vulnerable to simple manipulations, such as adding a ``teacher:'' prefix, which artificially increased system performance in the BEA2023 shared task \citep{tack-etal-2023-bea}, raising concerns about metric reliability. Adversarial robustness is an understudied aspect of dialogue system evaluation, but measuring the robustness of reference-free metrics can provide deeper insights into these vulnerabilities, helping to ensure their reliability across diverse scenarios.

Effective metrics should prioritize responses that are engaging, relevant, and grammatically correct, while penalizing generic or nonsensical ones. However, alignment with human judgment alone may not fully reflect a metric's ability to assess dialogue response quality. Metrics that perform well on human-alignment benchmarks might still be vulnerable to adversarial manipulations or fail to generalize across different datasets.
To address this, we analyzed metric performance on two datasets: one grounded and one ungrounded. Grounded dialogues incorporate external knowledge, such as facts, into responses, while ungrounded dialogues rely solely on conversational context. Using one of each dataset allows us to evaluate metrics in diverse conversational scenarios and assess their adaptability.

Our robustness benchmark directly tests resistance to adversarial manipulations at an instance level.
Unlike previous work that reported average scores for adversarial attacks, our approach quantifies how often metrics rank adversarial responses higher than the reference, providing granular insights into their vulnerabilities.  We generate 20 adversarial responses per conversation grouped into four types of attack: speaker tags, static responses, ungrammatical corruptions, and context repetition. The resulting benchmark reveals which metrics can reliably rank reference responses above adversarial ones.

We evaluated widely used metrics, including \dialogrpt, \uni, and our own implementation of an LLM prompt-based evaluation, \prompteval. The metrics were assessed based on their correlation with human annotations across multiple aspects of dialogue response quality---content, naturalness, relevance, and groundedness (when applicable)---as well as their robustness to adversarial responses. The results reveal variability in metrics' susceptibility to adversarial responses, with attacks based on the ungrounded dataset, DailyDialog, proving more difficult than those based on the grounded dataset, TopicalChat.

\section{Related Work}

\paragraph{Reference-Free Metrics} Reference-free evaluation metrics address the limitations of reference-based methods by assessing dialogue quality without relying on predefined responses. Examples include UniEval \citep{zhong-etal-2022-towards}, which evaluates multiple dialogue aspects using yes/no questions; DialogRPT \citep{gao-etal-2020-dialogue}, a series of models fine-tuned on Reddit data to assess relevance and engagement; FED \citep{mehri-eskenazi-2020-unsupervised}, which uses DialoGPT to evaluate turn-level dialogue qualities; and CPMI \citep{ren-etal-2023-c}, a metric based on pointwise mutual information. More recently, prompting LLMs has gained traction, with methods such as GPTScore \citep{fu2023gptscore} and G-Eval \citep{liu-etal-2023-g} leveraging token probabilities to generate weighted scores. In contrast, \citet{wang-etal-2023-chatgpt} employed a direct, unweighted assessment approach that assigns discrete scores without leveraging token probabilities.

\paragraph{Adversarial Robustness}
The robustness of dialogue evaluation metrics against adversarial responses has been explored in prior research, though it remains underemphasized compared to metrics' alignment with human judgment.
\citet{li-etal-2017-adversarial} introduced the Evaluator Reliability Error score, which quantifies how an evaluator's predictions deviate from gold standard accuracy across different scenarios. 
\citet{sai2019reevaluating} examined the Automatic Dialogue Evaluation Model (ADEM) \citep{lowe-etal-2017-towards}, which aimed to score dialogue responses on a scale of 1-5 by leveraging embeddings of the dialogue context, model response, and reference response. Although ADEM claimed to produce human-correlated scores, \citet{sai2019reevaluating} revealed discrepancies through adversarial tests, showing that ADEM failed to provide appropriate scores for multiple adversarial responses. \citet{sai2019reevaluating} demonstrated that robustness is a distinct aspect of meta-evaluation, separate from correlation with human judgment. However, despite their findings, adversarial robustness is not routinely tested on newly proposed metrics, leaving a critical gap in evaluation practices.

Building on \citet{sai2019reevaluating}, our work incorporates additional adversarial types, including nonsensical static responses inspired by \citet{baladon-etal-2023-retuyt} and fact repetition for grounded dialogue evaluation. Additionally, drawing on insights from \citet{hicke-etal-2023-assessing}, we introduce speaker tag attacks. Unlike \citet{sai2019reevaluating}, who reported average scores for original and corrupted responses, we analyze how often corrupted responses score higher than gold standard responses. This granular approach is particularly relevant for selecting metrics in reranking systems, where individual rankings directly influence system behavior.

\section{Method}

We tested different evaluation metrics by examining their correlation with human evaluation and their robustness against adversarial attacks.

\subsection{Adversarial Attack Categories}

The four categories of adversarial attacks are as follows:

\paragraph{Speaker Tags} We prepend responses with speaker tags such as "teacher:" or "user:" to test vulnerabilities similar to those observed in \dialogrpt \citep{hicke-etal-2023-assessing}. Although this vulnerability has only been demonstrated in \dialogrpt, the inclusion of conversational markers such as speaker tags could influence other metrics that rely on surface-level lexical patterns or conversational structure.

\paragraph{Static Responses} Static responses include fixed phrases (e.g., "Hello"), contextually inappropriate but conversationally engaging phrases (e.g., "I don't know, what do you think?"), and ungrammatical utterances (e.g., "I will do"). These attacks exploit potential biases in evaluation metrics that prioritize grammaticality or perceived conversational engagement over contextual relevance. For example, a metric might rate a grammatically correct but irrelevant static response higher than a human-written reference response if it fails to account for the lack of meaningful content. Testing these attacks helps determine whether metrics can distinguish between shallow engagement and genuine relevance to the dialogue history.

\paragraph{Ungrammatical Responses} Following \citet{sai2019reevaluating}, we generate ungrammatical responses by removing punctuation, omitting stopwords, retaining only nouns, altering token order, and randomly repeating words with a 0.2 probability. These modifications are designed to break the linguistic coherence of the response while retaining some surface-level content. This attack tests whether submetrics such as naturalness and grammar are robust enough to penalize responses with syntactic and grammatical errors. Evaluation systems that fail to properly penalize ungrammatical responses risk prioritizing content at the expense of linguistic quality, which is critical in dialogue systems.

\begin{table}
\small
\begin{center}
\begin{tabular}{p{0.5\textwidth}}
\textbf{Dialogue History:}\\
My throat is really dry.\\
Do you want to go get something to drink?\\
Yes, I'm parched.\\
What did you want to drink?\\
\\
\textbf{Reference Response:} \\
I was thinking about getting a soda.
\\
\end{tabular}
\caption{An example conversation from the DailyDialog Subset~\cite{li-etal-2017-dailydialog}. Each line of the dialogue history is a new speaker turn, and the reference response is used as the seed for adversarial responses.}
 \label{tab:dd_example}
\end{center}
\end{table}

\paragraph{Context Repetitions} We test three types of context-related attacks: repeating the last utterance of the dialogue history as the response, prepending that utterance to the reference response, and repeating a fact as the response (in grounded dialogues). These attacks specifically target relevance submetrics by exploiting the potential reliance of models on lexical overlap. Since evaluation systems often use random responses from unrelated dialogues as negative samples during training, repeated utterances from the same conversation may be incorrectly judged as highly relevant. For grounded dialogues, the fact repetition attack examines whether metrics can identify the lack of conversational coherence even when the response contains relevant factual information. By testing these scenarios, we assess whether metrics can distinguish between genuine relevance and superficial repetition.

\paragraph{}
For an example from DailyDialog, see the context and ground truth answer in Table \ref{tab:dd_example}.  Examples of each of the adversarial responses to this sample dialogue are listed in Table \ref{tab:dd_example_attacks}.  As fact repetition is only applicable in the grounded dataset, we present an example from that dataset.

\begin{table}
\small
\begin{center}
    \begin{tabular}{p{2cm}p{4.5cm}}
        \hline
       \multicolumn{2}{l}{\textit{Speaker Tag}}\\
       \hline
       ``teacher:'' pre. & \textcolor{mygreen}{teacher:} I was thinking about getting a soda.\\
       ``agent:'' prefix & \textcolor{mygreen}{agent:} I was thinking about getting a soda.\\
       ``user:'' prefix & \textcolor{mygreen}{user:} I was thinking about getting a soda.\\
       \hline
       \multicolumn{2}{l}{\textit{Static Responses}}\\
       \hline
       greeting & \textcolor{mygreen}{Hello} \\
       generic & \textcolor{mygreen}{I don't know} \\
         + question& \textcolor{mygreen}{I don't know, what do you think?} \\
         + repetition& \textcolor{mygreen}{I don't know, what do you think? I think} \\
       ungram. relevant & \textcolor{mygreen}{I'm sorry, can you repeat} \\
       ungram. relevant & \textcolor{mygreen}{I will do} \\
       gram. irrelevant & \textcolor{mygreen}{fantastic! how are you?} \\
       \hline
       \multicolumn{2}{l}{\textit{Ungrammatical Responses}}\\
       \hline
       No punctuation & I was thinking about getting a soda\textcolor{myred}{\sout{.}}\\
       No stopwords & I \textcolor{myred}{\sout{was}} thinking \textcolor{myred}{\sout{about}} getting \textcolor{myred}{\sout{a}} soda \textcolor{myred}{\sout{.}}\\
       Nouns \& verbs & I was thinking \textcolor{myred}{\sout{about}} getting \textcolor{myred}{\sout{a}} soda\textcolor{myred}{\sout{.}}\\
       Only nouns & \textcolor{myred}{\sout{I was thinking about getting a}} soda\textcolor{myred}{\sout{.}}\\
       Jumbled words & \textcolor{myyellow}{a I soda was about thinking getting .}\\
       Reversed words & \textcolor{myyellow}{. soda a getting about thinking was I}\\
       Repeated words & I \textcolor{mygreen}{I} was \textcolor{mygreen}{was} thinking about \textcolor{mygreen}{about} getting a soda.\\
       \hline
       \multicolumn{2}{l}{\textit{Context Repetition}}\\
       \hline
       Prev. utterance & \textcolor{mygreen}{What did you want to drink?}\\
       + reference & \textcolor{mygreen}{What did you want to drink?} I was thinking about getting a soda.\\
       Fact repetition & \textcolor{mygreen}{According to Canadian law, all radios are required to have at least 40\% of the music played be Canadian.} \\
       \hline
    \end{tabular}
    \caption{Adversarial responses for the example dialogue in Table \ref{tab:dd_example}. The fact repetition example shown is from the TopicalChat subset~\citet{mehri-eskenazi-2020-usr}. Additions to the reference sentence are shown in \textcolor{mygreen}{green}, deletions in \textcolor{myred}{\sout{red}}, and reorderings in \textcolor{myyellow}{yellow}.  Static responses always replace the reference response.}
 \label{tab:dd_example_attacks}
\end{center}
\end{table}

\subsection{Evaluation Datasets}
Two datasets were selected for their annotated candidate responses, allowing us to evaluate metrics for both human judgment correlation and robustness to adversarial responses

\paragraph{DailyDialog Subset}
For ungrounded dialogue data, we selected a subset of DailyDialog~\citep{li-etal-2017-dailydialog} released by \citet{zhao-etal-2020-designing}\footnote{\url{https://github.com/ZHAOTING/dialog-processing}}.  DailyDialog was chosen for its diverse and casual conversations, while the subset released by \citet{zhao-etal-2020-designing} was specifically selected for its detailed human annotations. This subset comprises 100 conversations, each with nine possible responses: the ground-truth response and eight additional synthetic responses. These responses have been annotated with human ratings for content, grammaticality, relevance, and overall quality.

\paragraph{Topical-Chat Subset}
For the grounded data, we used a subset of Topical-Chat released by \citet{mehri-eskenazi-2020-usr}, which extends the original Topical-Chat dataset \citep{gopalakrishnan2019topical}. The original Topical-Chat dataset comprises conversations generated by crowdworkers who incorporated provided facts into their turns. The subset released by \citet{mehri-eskenazi-2020-usr} builds on this by introducing additional candidate responses generated using various decoding strategies, along with human-written responses.
This subset includes 60 conversations, each with five candidate responses generated using decoding strategies and a second human-written response. These responses have been annotated with human ratings for naturalness, coherence, interestingness, groundedness, understandability, and overall quality.

\subsection{Adversarial Benchmark Creation}
We automatically generated adversarial responses for the DailyDialog and Topical-Chat subsets using predefined templates and rules, ensuring consistency and scalability. The benchmark construction process is applicable to any dialogue dataset without modification. Speaker tags were prepended to reference responses using fixed patterns, static responses were drawn from a curated list of phrases, and ungrammatical responses were generated programmatically by removing, repeating, or scrambling tokens. This automatic process enables the efficient creation of adversarial benchmarks without requiring additional human annotation.

\subsection{Adversarial Benchmark Evaluation}
We evaluate metrics along two axes: alignment with human judgment (via Kendall’s tau) and robustness to adversarial attacks (via attack success rate), investigating whether adversarial testing provides insights beyond those captured by human judgment evaluations.

The adversarial benchmark tests a metric's ability to correctly rank reference responses higher than adversarial ones. For each adversarial response, we calculate the success rate, defined as the percentage of instances where the reference response scores higher than the adversarial response. A score of 1 indicates perfect robustness against that attack, while a score of 0 means the attack is always rated higher than the reference.

By comparing results across these two evaluation dimensions—correlation with human judgment and adversarial robustness—we explore whether the adversarial benchmark reveals weaknesses or patterns not evident from correlation alone.

\begin{table*}[h]
\begin{center}
\small
\begin{tabular}{l|cc|cc|cc|cc|cc|cc|}
      & \multicolumn{2}{c|}{Content} & \multicolumn{2}{c|}{Naturalness} & \multicolumn{2}{c|}{Relevance} & \multicolumn{2}{c|}{Groundedness} & \multicolumn{2}{c|}{Overall}\\
    \textbf{Metrics} & DD & TC & DD & TC & DD & TC & DD & TC & DD & TC\\
    \hline
    \dialogrpt & \textit{-0.008} & 0.156 & -0.082 &  \textit{0.120} &  0.162 & \textit{0.075} & - & - & \textit{0.016} & 0.184\\

    \hline
    \uni & 0.322 & 0.459 & 0.117 & 0.374 & 0.381 & 0.466 & - & 0.452 & 0.349 & 0.487 \\
    \hline
    \llama-direct & 0.212 & 0.456 & 0.488  & 0.436 & 0.370 & 0.347 & - & 0.388 & 0.418 & 0.445 \\
    \llama-weighted & 0.227 & 0.464 & 0.365 & 0.430 & 0.476 & 0.406 & - & 0.389 & 0.393 & 0.460 \\
    \hline
    \mix-direct & 0.318 & 0.503 & 0.463 & 0.421 & 0.575 & 0.409 & - & 0.258 & 0.514 & 0.440 \\
    \mix-weighted & 0.278 & 0.479 & 0.251 & 0.373 & 0.509 & 0.394 & - & 0.272 & 0.488 & 0.452 \\ 
    \hline
    \gptthree-direct & 0.362 & 0.301 & 0.456 & 0.453 & 0.578 & 0.476 & - & 0.321 & 0.508 & 0.473 \\
    \gptthree-weighted & 0.318 & 0.363 & 0.405 & 0.431 & 0.532 & 0.441 & - & 0.319 & 0.488 & 0.474 \\
    \hline
    \gptfour-direct & \textbf{0.437} & \textbf{0.612} & \textbf{0.505} & \textbf{0.566} & \textbf{0.627} & 0.553 & - & \textbf{0.533} & \textbf{0.562} & \textbf{0.616} \\
    \gptfour-weighted & 0.392 & 0.592 & 0.458 & 0.504 & 0.613 & \textbf{0.572} & - & 0.132 & 0.526 & 0.590 \\

\end{tabular}
\caption{Turn-level Kendall's $\tau$ correlations with human judgement of different metrics on the DailyDialog subset (DD) and the  Topical-Chat Corpus (TC). \textit{Italicized} values are not statistically significant ($p>0.05$). The highest value in each column is bolded.  DailyDialog is not a grounded dataset, so groundedness was not tested.  DialogRPT is not able to evaluate for groundedness.}
 \label{tab:correlation}

 \end{center}
\end{table*}

\subsection{Dialogue Response Metrics}
\paragraph{DialogRPT}
\dialogrpt comprises five submetrics fine-tuned on Reddit data: likelihood of upvotes (updown $S_{u}$), user engagement (width $S_{w}$), discussion length (depth $S_{d}$), response relevance (human-vs-random $S_{hvr}$), and response naturalness (human-vs-machine $S_{hvm}$). These submetrics are combined using the following equation from the original implementation of \dialogrpt:
\begin{multline}
    S_{dialogRPT} = (S_{u} + 0.48 S_{d} - 0.5 S_{w}) \times\\
    0.5(S_{hvr} + S_{hvm})
\end{multline}
For the correlation experiment, we use $S_{dialogRPT}$ as the \texttt{overall} score. We use the weighted combination of updown, width, and depth 
as the \texttt{content} score.
We use human-vs-machine for \texttt{naturalness}, and human-vs-random for \texttt{relevance}.

\paragraph{UniEval}

\uni~\citep{zhong-etal-2022-towards} is a T5-based model trained to evaluate dialogue system responses on five aspects by answering yes/no questions. It was fine-tuned on synthetic data generated from 30k samples of the Topical-Chat dataset~\citep{gopalakrishnan2019topical}. For this study, we verified that none of the TopicalChat Subset used for evaluation overlaps with the training data used to fine-tune \uni.

For the grounded Topical-Chat dataset, we used \uni as originally designed, utilizing all five submetrics (\texttt{content}, \texttt{relevance}, \texttt{grammar}, \texttt{coherence}, and \texttt{groundedness}). The corresponding prompts were taken directly from the official \uni GitHub repository\footnote{\url{https://github.com/maszhongming/UniEval}}.

For the ungrounded DailyDialog dataset, modifications were necessary because it lacks the additional fact context present in Topical-Chat. We made zero-shot adjustments to the \uni prompts to better align with the dataset’s structure. Specifically, we removed references to the fact from the \texttt{content} prompt and used the original coherence prompt for \texttt{relevance}. Since groundedness evaluates whether a candidate response appropriately incorporates the fact, this submetric was deemed incompatible with DailyDialog and omitted. Similarly, \texttt{understandability}, which is included in the code but not the original \uni paper, was excluded. Detailed prompt adjustments are provided in Appendix \ref{sec:unieval}.

To compute the composite score, we weighted the submetrics to emphasize \texttt{content} and \texttt{relevance}, ensuring \texttt{grammar} was not disproportionately represented. The composite score was calculated as $0.4\cdot\text{content} + 0.2\cdot\text{grammar} + 0.4\cdot\text{relevance}$. This choice was motivated by a desire to remain faithful to the original metric while adapting it to the ungrounded dataset. The weighting was not optimized on a separate validation set but instead derived from an ablation study detailed in Appendix \ref{sec:unieval} (Table 6). Our goal was to preserve the relative importance of the submetrics as intended in the original design, while accommodating the limitations of the DailyDialog dataset.

\paragraph{PromptEval}
Several recent studies have used GPT-3.5 \citep{brown2020languagemodelsfewshotlearners} or GPT-4 \citep{openai2024gpt4technicalreport} to evaluate natural language generation by incorporating weighted scores based on probabilities. This method, seen in works such as GPTScore \citep{fu2023gptscore} and G-Eval \citep{liu-etal-2023-g}, uses token log probabilities to assign scores according to the likelihood of each model response. In our analysis, we compared two scoring methods: weighted scores, which adjust the evaluation based on the probabilities assigned to each response, and direct scores, which are direct assessments without weighting.

In the weighted variation, scores for each submetric are calculated as a weighted average of values from 1 to 5. The weight assigned to each value corresponds to the token probability of that value being generated in the response. This process is formally expressed as:

\[
S_{\text{submetric}} = \sum_{v=1}^{5} P(v) \cdot v
\]

where \(S_{\text{submetric}}\) represents the weighted score for the submetric, \(v\) denotes the possible values (1 through 5), and \(P(v)\) is the token probability of generating \(v\).

In addition to GPT-3.5 and GPT-4, we also tested Llama2 \citep{touvron2023llama2openfoundation} and Mixtral \citep{jiang2024mixtralexperts} as base LLMs for \prompteval.

We evaluated different submetrics on both grounded and ungrounded dialogue data. For the ungrounded DailyDialog subset, we used the following submetrics: content, grammar, and relevance. For the grounded Topical-Chat subset, we used content, naturalness, relevance, and groundedness to reflect the annotations of the dataset.

Our approach also includes an ``overall'' score prompt. We found that including this ``overall'' score and averaging it with the submetrics resulted in the highest correlation. The complete prompt is provided in Appendix \ref{sec:gpt_appendix}.


In the following tables, we refer to \prompteval metrics using the prefix \texttt{PE}. \gptfour represents the \prompteval built on top of 
\texttt{gpt-4o-2024-05-13}, and \gptthree built on top of \texttt{gpt-3.5-turbo-0125}.
Additional results from other checkpoints of GPT-3.5/4 are reported in Appendix \ref{sec:gpt_results_app}. \llama uses \texttt{Llama-2-70b-chat} and \mix uses \texttt{Mixtral-8x7B-Instruct}.

\begin{table*}[ht]
\begin{center}
\small
\begin{tabular}{l|cc|cc|cc|cc|cc|}
      & \multicolumn{2}{c|}{Speaker Tags} & \multicolumn{2}{c|}{Static Resp.} & \multicolumn{2}{c|}{Ungrammatical} & \multicolumn{2}{c|}{Context Rep.} & \multicolumn{2}{c|}{Avg}\\
    \textbf{Metrics} & DD & TC & DD & TC & DD & TC & DD & TC & DD & TC\\
    \hline
    \uni & 0.07 & \textbf{0.00} & 0.06 & \textbf{0.00} & 0.03 & \textbf{0.00} & 0.76 & 0.04 & 0.23 & \textbf{0.01} \\
    \hline
    \llama-direct & 
    0.32 & 0.05 &
    0.10 & 0.14 &
    0.11 & 0.03 & 
    0.40 & 0.20 & 
    0.23 & 0.10\\
    \llama-weighted &
    0.18 & 0.04 &
    0.08 & 0.10 &
    0.06 & 0.02 &
    0.30 & 0.21 &
    0.15 & 0.09\\
    \hline
    \mix-direct & 0.32 & 0.03 & 0.06 & 0.08 & 0.10 & 0.02 & 0.43 & 0.09 & 0.23 & 0.06\\
    \mix-weighted & 0.19 & 0.01 & 0.06 & 0.04 & 0.06 & \textbf{0.00} & 0.35 & 0.08 & 0.17 & 0.03\\ 
    \hline
    \gptthree-direct & 0.11 & 0.05 & 0.04 & 0.09 & 0.06 & 0.01 & 0.48 & 0.40 & 0.17 & 0.14 \\
    \gptthree-weighted & \textbf{0.01} & \textbf{0.00} & 0.04 & 0.05 & 0.03 & \textbf{0.00} & 0.39 & 0.09 & 0.12 & 0.03 \\
    \hline
    \gptfour-direct & 0.27 & 0.01 & \textbf{0.02} & 0.01 & 0.05 & \textbf{0.00} & 0.06 & \textbf{0.02} & 0.10 & \textbf{0.01} \\
    \gptfour-weighted & 0.03 & \textbf{0.00} & 0.03 & 0.01 & \textbf{0.01} & \textbf{0.00} & \textbf{0.05} & 0.04 & \textbf{0.03} & \textbf{0.01} \\

\end{tabular}
\caption{Attack success rate by attack category on DailyDialog and Topical-Chat.  The best performing system (lowest attack success rate) is bolded for each column.}
 \label{tab:grouped_attacks}

 \end{center}
\end{table*}

\section{Results}

\subsection{Correlation with Human Evaluation}

The results of our correlation experiment with the ungrounded data (Table \ref{tab:correlation}) show that \gptfour, using direct scoring, had the highest correlation in most categories. Interestingly, \gptthree performed better with individual prompts for each aspect. 
\mix and \llama were not far behind \gptthree in the \texttt{overall} score. 
\uni outperformed many \prompteval on the \texttt{content} correlation.

The correlation of \dialogrpt's content calculation with the annotated content scores lacked statistical significance, suggesting that it struggles to effectively measure response quality in terms of relevance and informativeness. This limitation may stem from a failure to generalize across diverse contexts due to its training conditions. Specifically, \dialogrpt was fine-tuned on Reddit data, which may not adequately reflect the full range of dialogue scenarios present in other datasets. Furthermore, its use of upvotes, user engagement, and discussion length as proxies for dialogue quality may not capture the qualitative aspects that humans evaluators prioritize, such as coherence, relevance, and engagement.

Given \dialogrpt's low correlation with human ratings and unsuitability for grounded dialogues, further experiments will focus on \uni and \prompteval. These methods have shown more promise in aligning with human evaluations and offer better adaptability.

\uni performs more competitively on the grounded dataset than on the ungrounded one. This may be due to being trained on pseudo-data seeded with conversations from Topical-Chat. Although \gptfour still outperforms \uni on the grounded dataset, \uni demonstrates impressive performance given its size. At approximately 800M parameters, it is the smallest model we tested, yet it often achieved better correlation with human judgment than significantly larger models such as the 7B Llama and the 8x7B Mixtral.

When comparing the weighted and direct versions of \prompteval, the direct version consistently performs slightly better in both DailyDialog and Topical-Chat. Although the differences are minimal, the direct version of \gptfour emerges as the top performer, with the weighted version as a close second.

The correlations for submetric evaluations were similar between DailyDialog and Topical-Chat, with models that performed well on one dataset generally performing well on the other. Although there were some variations in the submetric scores, the overall correlations remained consistent between both data sets, with the exception that \uni scored higher in content and groundedness than \mix or \llama, but significantly lower in naturalness.

\subsection{Vulnerability Against Adversarial Attacks}
We tested the robustness of the metrics with our proposed benchmark. Table \ref{tab:grouped_attacks} shows the models' vulnerability scores averaged over each attack category on DailyDialog and Topical-Chat.  The higher the number, the more susceptible the metric is to attacks in that category.

Built on the comparatively older T5 model, \uni performs surprisingly well in terms of human correlation and robustness, particularly on the Topical-Chat subset, where it ranks just behind \gptfour.
However, on the DailyDialog dataset, \uni is vulnerable to context repetition attacks, often scoring perfect dialogue history copies highly revealing a tendency to reward similarity between context and response. This issue is less pronounced in the TopicalChat dataset, where \uni handles such attacks more effectively. However, \gptfour consistently outperforms \uni by penalizing context repetition on both datasets. Ideally, relevance or content metrics should penalize redundant responses regardless of the dataset.

Both \mix and \gptfour-direct are particularly vulnerable to speaker tag attacks, potentially due to misinterpretation of these tags as genuine speaker changes. 
Although \gptfour-direct had the highest correlation with human judgment, its frequent ties made it more susceptible to these speaker tag attacks. 
In contrast, the weighted versions of the \prompteval models, including the weighted \gptfour, demonstrated greater resilience across the board, showing that the weighting methods effectively reduce vulnerabilities by better handling of the links and improving robustness.

Although all models performed well overall against static and ungrammatical responses, the \prompteval metrics occasionally failed to distinguish between unpunctuated and correctly punctuated responses, treating them as equivalent in a sizable percentage of cases (21--39\%). This indicates a potential oversight in how punctuation errors are treated in the overall score, since the models prioritize high-level conversational flow over grammatical accuracy.

In general, our findings suggest that metrics with high correlation to human judgment are not uniformly robust across all types of attacks. While \gptfour-weighted demonstrates resilience across datasets, metrics like \uni and \mix perform well in certain areas but exhibit vulnerabilities to specific attacks, such as context repetition and speaker tag manipulations. These results highlight the value of evaluating both correlation with human judgment and robustness against adversarial attacks to gain a more comprehensive understanding of dialogue evaluation metrics.

\subsection{Dataset-Level Robustness Differences}

To better understand the robustness of the evaluation metrics, we compared adversarial responses made from two different datasets. This comparison reveals that models respond differently to adversarial attacks depending on the dataset.

Our analysis of static responses revealed some variability across the datasets. While \gptfour maintains low static attack success rates across both DailyDialog and Topical-Chat, \llama and \mix exhibit increased susceptibility on Topical-Chat. In contrast, \uni shows a notable reduction in static attack success rates, dropping to 0.00 in Topical-Chat.

There was a slight improvement in robustness against speaker tags and ungrammatical responses when moving from DailyDialog to Topical-Chat. This indicates that the models are somewhat more resilient to these types of attacks in the Topical-Chat dataset.

\begin{table}[t]
\small
    \centering
    \begin{tabular}{c|c}
       Dataset  & Avg response length \\
       \hline
       DailyDialog & 7.8 \\
       Topical-Chat & 22.9 \\
    \end{tabular}
    \caption{Average number of tokens in the ground truth response for each dataset.}
    \label{tab:response_length}
\end{table}

In particular, some models (\gptthree-weighted, both \mix and \uni) showed a decrease in successful context repetition attacks from DailyDialog to Topical-Chat. We theorize that this is due to the longer conversations and responses in Topical-Chat compared to DailyDialog. As shown in Table \ref{tab:response_length}, the average number of tokens in a Topical-Chat response is higher than in a DailyDialog response.
Longer responses may help models better distinguish between relevant and irrelevant content, thereby reducing the effectiveness of context repetition attacks.

The differences between DailyDialog and TopicalChat demonstrate the impact of the characteristics of the dataset on the robustness of the evaluation. DailyDialog, with its shorter and simpler conversations, tends to expose weaknesses in models such as \uni, which can be fooled by repeated context. In contrast, the longer and more grounded nature of TopicalChat offers a more challenging environment for models to maintain consistency in evaluating relevance and content, where the added complexity can help models such as \gptfour better distinguish meaningful responses from manipulative attacks.


\begin{figure*}[t]
\begin{center}
\hspace{-0.1\textwidth}
\includegraphics[width=1.1\textwidth]{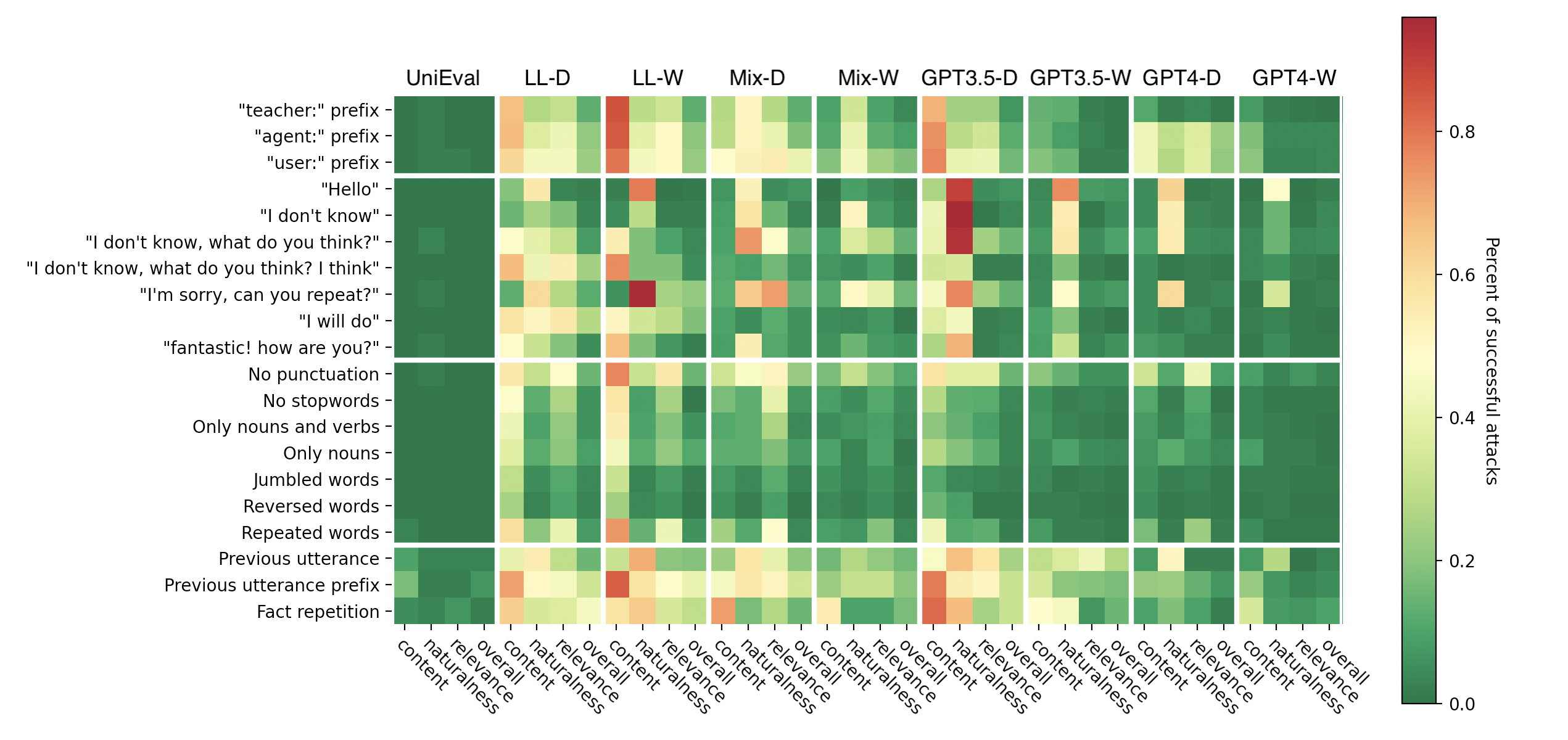} 
\caption{The effectiveness of each attack type on different submetrics on the combined dataset of DailyDialog and Topical-Chat.}
\label{fig:master}
\end{center}
\end{figure*}

\subsection{Submetric Performance on Adversarial Benchmarks}

To better understand how the models judge responses, we analyzed the performance of each submetric against the adversarial attacks of our benchmark. Figure~\ref{fig:master} provides a heatmap illustrates the robustness of each submetric across attack types, with results averaged over the DailyDialog and Topical-Chat datasets. This analysis evaluates whether the metrics align with our expectations for each submetric’s behavior under different attack types.

\paragraph{Speaker Tag Attacks}
We hypothesized that speaker tag attacks would lower the \texttt{naturalness} score, as it is unnatural for a reply to begin with a tag like "user:". Surprisingly, we did not observe a strong difference in \texttt{naturalness} scores between tagged and untagged responses from most of the metrics (except \uni and \gptfour-W). Instead, some metrics (e.g., \llama-W, \gptthree-D) assigned a higher \texttt{content} score to the tagged response, artificially increasing the overall score in a manner reminiscent of the behavior observed from DialogRPT in \citet{hicke-etal-2023-assessing}. This indicates that some models may misinterpret speaker tags as adding meaningful information.

\paragraph{Static Responses}
Static responses, such as "I don't know," were expected to score lower on \texttt{relevance}, as they lack context-specific information. This expectation held true for \gptthree, \gptfour, and \uni, but was less consistent for \llama. The \texttt{content} score of the static responses was reliably lower than that of the reference response with the same metrics. The grammatically correct static responses scored high on \texttt{naturalness}, which aligns with the expectation that those grammatically correct yet uninformative responses might still appear "natural."

\paragraph{Ungrammatical Responses}
We anticipated that \texttt{naturalness} would be robust against ungrammatical responses, and this was confirmed in all metrics. However, responses without punctuation were scored similarly to responses with punctuation in all \prompteval metrics except \gptfour. This indicates a surprising tolerance for surface-level errors in most of the models tested, potentially reflecting their focus on higher-level semantic coherence over grammatical precision. Interestingly, ungrammatical responses were also somewhat more likely to be rated lower on \texttt{relevance} and \texttt{content}, suggesting that these errors negatively influence the perceived meaning and utility of the response, in addition to its fluency.

\paragraph{Context Repetition}
These attacks were included to target the \texttt{relevance} submetric, exploiting its tendency to favor responses closely aligned with the dialogue history. While this vulnerability was evident in most metrics, we hoped that \texttt{content} would be robust enough to offset it. However, the \texttt{content} submetric was more vulnerable than anticipated to these attacks, particularly when the context was appended to the previous utterance. This suggests that many metrics fail to penalize repeated content adequately, potentially mistaking verbosity for meaningfulness.

\paragraph{}The varying performance of submetrics across attacks highlights both strengths and limitations in how models evaluate responses. Speaker tag attacks reveal an over-reliance on superficial indicators of content, while static responses expose gaps in distinguishing between naturalness and relevance. Context repetition attacks demonstrate the need for better safeguards against repetition bias.

\section{Discussion}
The adversarial benchmark applied in this study reveals vulnerabilities in dialogue evaluation metrics that are often overlooked when relying solely on human judgment correlation. By testing a variety of metrics, including \dialogrpt, \uni, and \prompteval, we uncovered distinctions that traditional evaluations do not capture.

For example, while both the weighted and direct versions of \gptfour performed similarly in terms of correlation with human judgment, the weighted version demonstrated greater robustness across attack types and data sets. This illustrates that metrics which appear equally effective in standard evaluations may differ significantly in their resilience to adversarial manipulations, underscoring the necessity of adversarial testing to fully assess their reliability. The improved ability of the weighted version to penalize manipulative strategies, such as speaker tag prefixes, makes it a more reliable metric for dialogue evaluation.

The results of our experiments demonstrate that employing a variety of adversarial attacks is essential to identify specific vulnerabilities in dialogue evaluation metrics. Each attack type exposed different weaknesses, underscoring the value of testing metrics against diverse manipulations.
Speaker tag manipulations were particularly challenging for direct versions of models evaluated through \prompteval, including \gptfour, raising concerns for real-world applications. 
In conversational systems, correctly interpreting speaker identity and ensuring that irrelevant tags do not influence response scoring are crucial for reliable evaluation. This is especially important in multi-turn dialogues, where tags like ``user'' or ``agent'' should not alter the perceived quality of the response.

Static responses serve as a sanity check, ensuring models do not over-reward grammatical correctness at the expense of relevance. This is particularly important for systems designed to encourage meaningful interactions.
Although most models handle ungrammatical responses well, including these tests ensures that metrics prioritize both content and form.

Finally, context repetition attacks test the ability of a model to detect redundancy. Repetition of previous statements adds no value, yet models like \llama and \mix often reward these responses with higher content scores. Evaluating the responses of the metrics to repeated context ensures that they prioritize informative and context-aware dialogue.

By incorporating various types of attack, our adversarial benchmark reveals weaknesses that traditional human judgment alignment may overlook. The results of our experiments demonstrate its usefulness in developing more robust and reliable dialogue evaluation metrics.

\section{Conclusion}

Our findings demonstrate the value of adversarial testing as a complement to traditional human judgment alignment. Although correlation with human ratings provides insight into how well metrics align with aspects such as content or relevance, adversarial testing reveals vulnerabilities that are not captured by human judgment alone. Together, these dimensions offer a more comprehensive understanding of metric performance.

By applying our adversarial benchmark, we uncovered notable differences in the robustness of metrics, particularly between the weighted and discrete versions of \gptfour, which exhibited similar correlations with human judgment but varied in their susceptibility to context repetition and speaker tag attacks.

Our results demonstrate the complementary nature of these evaluation dimensions: human judgment alignment assesses the quality of typical responses, while adversarial testing probes the boundaries of metric reliability under challenging conditions. This dual approach provides actionable insights for improving dialogue evaluation metrics and ensuring their adaptability across diverse scenarios.

Looking ahead, the flexibility of the adversarial benchmark opens opportunities for broader applications in text generation, such as summarization and machine translation. Developing targeted adversarial tests for these domains presents an exciting avenue for future research, fostering more reliable and robust evaluation practices across NLP tasks.

\section*{Limitations}
This adversarial benchmark was created programmatically using Python scripts applied to the DailyDialog and Topical-Chat datasets, reflecting the authors’ assumptions about adversarial vulnerabilities. Although the data generated was carefully designed, it may not comprehensively capture all potential weaknesses in evaluation metrics or represent the full range of real-world challenges. Therefore, caution should be exercised when generalizing our findings to unforeseen adversarial situations or datasets.

Our use of LLMs for evaluation introduces additional considerations, such as the influence of prompt design and the order of response presentation, which can impact the results. The prompt search for \prompteval was not exhaustive, and different prompt designs could lead to alternative outcomes. Understanding and mitigating these biases remains an important direction for future research.

It is also important to note that language models such as GPT-4 and Llama are continuously updated and improved. Consequently, the findings of our study are specific to the versions and configurations used during the research and may not directly apply to future iterations of these models.

Finally, while metrics' correlation with human judgment serves as a benchmark, human evaluation itself is subjective and can introduce biases. Developing objective methods to establish gold standards for dialogue evaluation remains an open challenge and an area of future exploration.

\bibliography{anthology,custom}

\begin{thebibliography}{22}
\providecommand{\natexlab}[1]{#1}

\bibitem[{Balad{\'o}n et~al.(2023)Balad{\'o}n, Sastre, Chiruzzo, and Ros{\'a}}]{baladon-etal-2023-retuyt}
Alexis Balad{\'o}n, Ignacio Sastre, Luis Chiruzzo, and Aiala Ros{\'a}. 2023.
\newblock \href {https://doi.org/10.18653/v1/2023.bea-1.61} {{RETUYT}-{I}n{C}o at {BEA} 2023 shared task: Tuning open-source {LLM}s for generating teacher responses}.
\newblock In \emph{Proceedings of the 18th Workshop on Innovative Use of NLP for Building Educational Applications (BEA 2023)}, pages 756--765, Toronto, Canada. Association for Computational Linguistics.

\bibitem[{Brown et~al.(2020)Brown, Mann, Ryder, Subbiah, Kaplan, Dhariwal, Neelakantan, Shyam, Sastry, Askell, Agarwal, Herbert-Voss, Krueger, Henighan, Child, Ramesh, Ziegler, Wu, Winter, Hesse, Chen, Sigler, Litwin, Gray, Chess, Clark, Berner, McCandlish, Radford, Sutskever, and Amodei}]{brown2020languagemodelsfewshotlearners}
Tom~B. Brown, Benjamin Mann, Nick Ryder, Melanie Subbiah, Jared Kaplan, Prafulla Dhariwal, Arvind Neelakantan, Pranav Shyam, Girish Sastry, Amanda Askell, Sandhini Agarwal, Ariel Herbert-Voss, Gretchen Krueger, Tom Henighan, Rewon Child, Aditya Ramesh, Daniel~M. Ziegler, Jeffrey Wu, Clemens Winter, Christopher Hesse, Mark Chen, Eric Sigler, Mateusz Litwin, Scott Gray, Benjamin Chess, Jack Clark, Christopher Berner, Sam McCandlish, Alec Radford, Ilya Sutskever, and Dario Amodei. 2020.
\newblock \href {https://arxiv.org/abs/2005.14165} {Language models are few-shot learners}.
\newblock \emph{Preprint}, arXiv:2005.14165.

\bibitem[{Fu et~al.(2023)Fu, Ng, Jiang, and Liu}]{fu2023gptscore}
Jinlan Fu, See-Kiong Ng, Zhengbao Jiang, and Pengfei Liu. 2023.
\newblock Gptscore: Evaluate as you desire.
\newblock \emph{arXiv preprint arXiv:2302.04166}.

\bibitem[{Gao et~al.(2020)Gao, Zhang, Galley, Brockett, and Dolan}]{gao-etal-2020-dialogue}
Xiang Gao, Yizhe Zhang, Michel Galley, Chris Brockett, and Bill Dolan. 2020.
\newblock \href {https://doi.org/10.18653/v1/2020.emnlp-main.28} {Dialogue response ranking training with large-scale human feedback data}.
\newblock In \emph{Proceedings of the 2020 Conference on Empirical Methods in Natural Language Processing (EMNLP)}, pages 386--395, Online. Association for Computational Linguistics.

\bibitem[{Gopalakrishnan et~al.(2019)Gopalakrishnan, Hedayatnia, Chen, Gottardi, Kwatra, Venkatesh, Gabriel, and Hakkani-Tür}]{gopalakrishnan2019topical}
Karthik Gopalakrishnan, Behnam Hedayatnia, Qinlang Chen, Anna Gottardi, Sanjeev Kwatra, Anu Venkatesh, Raefer Gabriel, and Dilek Hakkani-Tür. 2019.
\newblock \href {https://doi.org/10.21437/Interspeech.2019-3079} {{Topical-Chat: Towards Knowledge-Grounded Open-Domain Conversations}}.
\newblock In \emph{Proc. Interspeech 2019}, pages 1891--1895.

\bibitem[{Hicke et~al.(2023)Hicke, Masand, Guo, and Gangavarapu}]{hicke-etal-2023-assessing}
Yann Hicke, Abhishek Masand, Wentao Guo, and Tushaar Gangavarapu. 2023.
\newblock \href {https://doi.org/10.18653/v1/2023.bea-1.60} {Assessing the efficacy of large language models in generating accurate teacher responses}.
\newblock In \emph{Proceedings of the 18th Workshop on Innovative Use of NLP for Building Educational Applications (BEA 2023)}, pages 745--755, Toronto, Canada. Association for Computational Linguistics.

\bibitem[{Jiang et~al.(2024)Jiang, Sablayrolles, Roux, Mensch, Savary, Bamford, Chaplot, de~las Casas, Hanna, Bressand, Lengyel, Bour, Lample, Lavaud, Saulnier, Lachaux, Stock, Subramanian, Yang, Antoniak, Scao, Gervet, Lavril, Wang, Lacroix, and Sayed}]{jiang2024mixtralexperts}
Albert~Q. Jiang, Alexandre Sablayrolles, Antoine Roux, Arthur Mensch, Blanche Savary, Chris Bamford, Devendra~Singh Chaplot, Diego de~las Casas, Emma~Bou Hanna, Florian Bressand, Gianna Lengyel, Guillaume Bour, Guillaume Lample, Lélio~Renard Lavaud, Lucile Saulnier, Marie-Anne Lachaux, Pierre Stock, Sandeep Subramanian, Sophia Yang, Szymon Antoniak, Teven~Le Scao, Théophile Gervet, Thibaut Lavril, Thomas Wang, Timothée Lacroix, and William~El Sayed. 2024.
\newblock \href {https://arxiv.org/abs/2401.04088} {Mixtral of experts}.
\newblock \emph{Preprint}, arXiv:2401.04088.

\bibitem[{Li et~al.(2017{\natexlab{a}})Li, Monroe, Shi, Jean, Ritter, and Jurafsky}]{li-etal-2017-adversarial}
Jiwei Li, Will Monroe, Tianlin Shi, S{\'e}bastien Jean, Alan Ritter, and Dan Jurafsky. 2017{\natexlab{a}}.
\newblock \href {https://doi.org/10.18653/v1/D17-1230} {Adversarial learning for neural dialogue generation}.
\newblock In \emph{Proceedings of the 2017 Conference on Empirical Methods in Natural Language Processing}, pages 2157--2169, Copenhagen, Denmark. Association for Computational Linguistics.

\bibitem[{Li et~al.(2017{\natexlab{b}})Li, Su, Shen, Li, Cao, and Niu}]{li-etal-2017-dailydialog}
Yanran Li, Hui Su, Xiaoyu Shen, Wenjie Li, Ziqiang Cao, and Shuzi Niu. 2017{\natexlab{b}}.
\newblock \href {https://aclanthology.org/I17-1099} {{D}aily{D}ialog: A manually labelled multi-turn dialogue dataset}.
\newblock In \emph{Proceedings of the Eighth International Joint Conference on Natural Language Processing (Volume 1: Long Papers)}, pages 986--995, Taipei, Taiwan. Asian Federation of Natural Language Processing.

\bibitem[{Liu et~al.(2016)Liu, Lowe, Serban, Noseworthy, Charlin, and Pineau}]{liu-etal-2016-evaluate}
Chia-Wei Liu, Ryan Lowe, Iulian Serban, Mike Noseworthy, Laurent Charlin, and Joelle Pineau. 2016.
\newblock \href {https://doi.org/10.18653/v1/D16-1230} {How {NOT} to evaluate your dialogue system: An empirical study of unsupervised evaluation metrics for dialogue response generation}.
\newblock In \emph{Proceedings of the 2016 Conference on Empirical Methods in Natural Language Processing}, pages 2122--2132, Austin, Texas. Association for Computational Linguistics.

\bibitem[{Liu et~al.(2023)Liu, Iter, Xu, Wang, Xu, and Zhu}]{liu-etal-2023-g}
Yang Liu, Dan Iter, Yichong Xu, Shuohang Wang, Ruochen Xu, and Chenguang Zhu. 2023.
\newblock \href {https://aclanthology.org/2023.emnlp-main.153} {{G}-eval: {NLG} evaluation using gpt-4 with better human alignment}.
\newblock In \emph{Proceedings of the 2023 Conference on Empirical Methods in Natural Language Processing}, pages 2511--2522, Singapore. Association for Computational Linguistics.

\bibitem[{Lowe et~al.(2017)Lowe, Noseworthy, Serban, Angelard-Gontier, Bengio, and Pineau}]{lowe-etal-2017-towards}
Ryan Lowe, Michael Noseworthy, Iulian~Vlad Serban, Nicolas Angelard-Gontier, Yoshua Bengio, and Joelle Pineau. 2017.
\newblock \href {https://doi.org/10.18653/v1/P17-1103} {Towards an automatic {T}uring test: Learning to evaluate dialogue responses}.
\newblock In \emph{Proceedings of the 55th Annual Meeting of the Association for Computational Linguistics (Volume 1: Long Papers)}, pages 1116--1126, Vancouver, Canada. Association for Computational Linguistics.

\bibitem[{Mehri and Eskenazi(2020{\natexlab{a}})}]{mehri-eskenazi-2020-unsupervised}
Shikib Mehri and Maxine Eskenazi. 2020{\natexlab{a}}.
\newblock \href {https://aclanthology.org/2020.sigdial-1.28} {Unsupervised evaluation of interactive dialog with {D}ialo{GPT}}.
\newblock In \emph{Proceedings of the 21th Annual Meeting of the Special Interest Group on Discourse and Dialogue}, pages 225--235, 1st virtual meeting. Association for Computational Linguistics.

\bibitem[{Mehri and Eskenazi(2020{\natexlab{b}})}]{mehri-eskenazi-2020-usr}
Shikib Mehri and Maxine Eskenazi. 2020{\natexlab{b}}.
\newblock \href {https://doi.org/10.18653/v1/2020.acl-main.64} {{USR}: An unsupervised and reference free evaluation metric for dialog generation}.
\newblock In \emph{Proceedings of the 58th Annual Meeting of the Association for Computational Linguistics}, pages 681--707, Online. Association for Computational Linguistics.

\bibitem[{OpenAI et~al.(2024)OpenAI, Achiam, Adler, Agarwal, Ahmad, Akkaya, Aleman, Almeida, Altenschmidt, Altman, Anadkat, Avila, Babuschkin, Balaji, Balcom, Baltescu, Bao, Bavarian, Belgum, Bello, Berdine, Bernadett-Shapiro, Berner, Bogdonoff, Boiko, Boyd, Brakman, Brockman, Brooks, Brundage, Button, Cai, Campbell, Cann, Carey, Carlson, Carmichael, Chan, Chang, Chantzis, Chen, Chen, Chen, Chen, Chen, Chess, Cho, Chu, Chung, Cummings, Currier, Dai, Decareaux, Degry, Deutsch, Deville, Dhar, Dohan, Dowling, Dunning, Ecoffet, Eleti, Eloundou, Farhi, Fedus, Felix, Fishman, Forte, Fulford, Gao, Georges, Gibson, Goel, Gogineni, Goh, Gontijo-Lopes, Gordon, Grafstein, Gray, Greene, Gross, Gu, Guo, Hallacy, Han, Harris, He, Heaton, Heidecke, Hesse, Hickey, Hickey, Hoeschele, Houghton, Hsu, Hu, Hu, Huizinga, Jain, Jain, Jang, Jiang, Jiang, Jin, Jin, Jomoto, Jonn, Jun, Kaftan, Łukasz Kaiser, Kamali, Kanitscheider, Keskar, Khan, Kilpatrick, Kim, Kim, Kim, Kirchner, Kiros, Knight, Kokotajlo, Łukasz Kondraciuk,
  Kondrich, Konstantinidis, Kosic, Krueger, Kuo, Lampe, Lan, Lee, Leike, Leung, Levy, Li, Lim, Lin, Lin, Litwin, Lopez, Lowe, Lue, Makanju, Malfacini, Manning, Markov, Markovski, Martin, Mayer, Mayne, McGrew, McKinney, McLeavey, McMillan, McNeil, Medina, Mehta, Menick, Metz, Mishchenko, Mishkin, Monaco, Morikawa, Mossing, Mu, Murati, Murk, Mély, Nair, Nakano, Nayak, Neelakantan, Ngo, Noh, Ouyang, O'Keefe, Pachocki, Paino, Palermo, Pantuliano, Parascandolo, Parish, Parparita, Passos, Pavlov, Peng, Perelman, de~Avila Belbute~Peres, Petrov, de~Oliveira~Pinto, Michael, Pokorny, Pokrass, Pong, Powell, Power, Power, Proehl, Puri, Radford, Rae, Ramesh, Raymond, Real, Rimbach, Ross, Rotsted, Roussez, Ryder, Saltarelli, Sanders, Santurkar, Sastry, Schmidt, Schnurr, Schulman, Selsam, Sheppard, Sherbakov, Shieh, Shoker, Shyam, Sidor, Sigler, Simens, Sitkin, Slama, Sohl, Sokolowsky, Song, Staudacher, Such, Summers, Sutskever, Tang, Tezak, Thompson, Tillet, Tootoonchian, Tseng, Tuggle, Turley, Tworek, Uribe, Vallone,
  Vijayvergiya, Voss, Wainwright, Wang, Wang, Wang, Ward, Wei, Weinmann, Welihinda, Welinder, Weng, Weng, Wiethoff, Willner, Winter, Wolrich, Wong, Workman, Wu, Wu, Wu, Xiao, Xu, Yoo, Yu, Yuan, Zaremba, Zellers, Zhang, Zhang, Zhao, Zheng, Zhuang, Zhuk, and Zoph}]{openai2024gpt4technicalreport}
OpenAI, Josh Achiam, Steven Adler, Sandhini Agarwal, Lama Ahmad, Ilge Akkaya, Florencia~Leoni Aleman, Diogo Almeida, Janko Altenschmidt, Sam Altman, Shyamal Anadkat, Red Avila, Igor Babuschkin, Suchir Balaji, Valerie Balcom, Paul Baltescu, Haiming Bao, Mohammad Bavarian, Jeff Belgum, Irwan Bello, Jake Berdine, Gabriel Bernadett-Shapiro, Christopher Berner, Lenny Bogdonoff, Oleg Boiko, Madelaine Boyd, Anna-Luisa Brakman, Greg Brockman, Tim Brooks, Miles Brundage, Kevin Button, Trevor Cai, Rosie Campbell, Andrew Cann, Brittany Carey, Chelsea Carlson, Rory Carmichael, Brooke Chan, Che Chang, Fotis Chantzis, Derek Chen, Sully Chen, Ruby Chen, Jason Chen, Mark Chen, Ben Chess, Chester Cho, Casey Chu, Hyung~Won Chung, Dave Cummings, Jeremiah Currier, Yunxing Dai, Cory Decareaux, Thomas Degry, Noah Deutsch, Damien Deville, Arka Dhar, David Dohan, Steve Dowling, Sheila Dunning, Adrien Ecoffet, Atty Eleti, Tyna Eloundou, David Farhi, Liam Fedus, Niko Felix, Simón~Posada Fishman, Juston Forte, Isabella Fulford, Leo
  Gao, Elie Georges, Christian Gibson, Vik Goel, Tarun Gogineni, Gabriel Goh, Rapha Gontijo-Lopes, Jonathan Gordon, Morgan Grafstein, Scott Gray, Ryan Greene, Joshua Gross, Shixiang~Shane Gu, Yufei Guo, Chris Hallacy, Jesse Han, Jeff Harris, Yuchen He, Mike Heaton, Johannes Heidecke, Chris Hesse, Alan Hickey, Wade Hickey, Peter Hoeschele, Brandon Houghton, Kenny Hsu, Shengli Hu, Xin Hu, Joost Huizinga, Shantanu Jain, Shawn Jain, Joanne Jang, Angela Jiang, Roger Jiang, Haozhun Jin, Denny Jin, Shino Jomoto, Billie Jonn, Heewoo Jun, Tomer Kaftan, Łukasz Kaiser, Ali Kamali, Ingmar Kanitscheider, Nitish~Shirish Keskar, Tabarak Khan, Logan Kilpatrick, Jong~Wook Kim, Christina Kim, Yongjik Kim, Jan~Hendrik Kirchner, Jamie Kiros, Matt Knight, Daniel Kokotajlo, Łukasz Kondraciuk, Andrew Kondrich, Aris Konstantinidis, Kyle Kosic, Gretchen Krueger, Vishal Kuo, Michael Lampe, Ikai Lan, Teddy Lee, Jan Leike, Jade Leung, Daniel Levy, Chak~Ming Li, Rachel Lim, Molly Lin, Stephanie Lin, Mateusz Litwin, Theresa Lopez, Ryan
  Lowe, Patricia Lue, Anna Makanju, Kim Malfacini, Sam Manning, Todor Markov, Yaniv Markovski, Bianca Martin, Katie Mayer, Andrew Mayne, Bob McGrew, Scott~Mayer McKinney, Christine McLeavey, Paul McMillan, Jake McNeil, David Medina, Aalok Mehta, Jacob Menick, Luke Metz, Andrey Mishchenko, Pamela Mishkin, Vinnie Monaco, Evan Morikawa, Daniel Mossing, Tong Mu, Mira Murati, Oleg Murk, David Mély, Ashvin Nair, Reiichiro Nakano, Rajeev Nayak, Arvind Neelakantan, Richard Ngo, Hyeonwoo Noh, Long Ouyang, Cullen O'Keefe, Jakub Pachocki, Alex Paino, Joe Palermo, Ashley Pantuliano, Giambattista Parascandolo, Joel Parish, Emy Parparita, Alex Passos, Mikhail Pavlov, Andrew Peng, Adam Perelman, Filipe de~Avila Belbute~Peres, Michael Petrov, Henrique~Ponde de~Oliveira~Pinto, Michael, Pokorny, Michelle Pokrass, Vitchyr~H. Pong, Tolly Powell, Alethea Power, Boris Power, Elizabeth Proehl, Raul Puri, Alec Radford, Jack Rae, Aditya Ramesh, Cameron Raymond, Francis Real, Kendra Rimbach, Carl Ross, Bob Rotsted, Henri Roussez,
  Nick Ryder, Mario Saltarelli, Ted Sanders, Shibani Santurkar, Girish Sastry, Heather Schmidt, David Schnurr, John Schulman, Daniel Selsam, Kyla Sheppard, Toki Sherbakov, Jessica Shieh, Sarah Shoker, Pranav Shyam, Szymon Sidor, Eric Sigler, Maddie Simens, Jordan Sitkin, Katarina Slama, Ian Sohl, Benjamin Sokolowsky, Yang Song, Natalie Staudacher, Felipe~Petroski Such, Natalie Summers, Ilya Sutskever, Jie Tang, Nikolas Tezak, Madeleine~B. Thompson, Phil Tillet, Amin Tootoonchian, Elizabeth Tseng, Preston Tuggle, Nick Turley, Jerry Tworek, Juan Felipe~Cerón Uribe, Andrea Vallone, Arun Vijayvergiya, Chelsea Voss, Carroll Wainwright, Justin~Jay Wang, Alvin Wang, Ben Wang, Jonathan Ward, Jason Wei, CJ~Weinmann, Akila Welihinda, Peter Welinder, Jiayi Weng, Lilian Weng, Matt Wiethoff, Dave Willner, Clemens Winter, Samuel Wolrich, Hannah Wong, Lauren Workman, Sherwin Wu, Jeff Wu, Michael Wu, Kai Xiao, Tao Xu, Sarah Yoo, Kevin Yu, Qiming Yuan, Wojciech Zaremba, Rowan Zellers, Chong Zhang, Marvin Zhang, Shengjia
  Zhao, Tianhao Zheng, Juntang Zhuang, William Zhuk, and Barret Zoph. 2024.
\newblock \href {https://arxiv.org/abs/2303.08774} {Gpt-4 technical report}.
\newblock \emph{Preprint}, arXiv:2303.08774.

\bibitem[{Ren et~al.(2023)Ren, Sidhu, Zeng, Gangi~Reddy, Ji, and Zhai}]{ren-etal-2023-c}
Liliang Ren, Mankeerat Sidhu, Qi~Zeng, Revanth Gangi~Reddy, Heng Ji, and ChengXiang Zhai. 2023.
\newblock \href {https://doi.org/10.18653/v1/2023.dialdoc-1.9} {{C}-{PMI}: Conditional pointwise mutual information for turn-level dialogue evaluation}.
\newblock In \emph{Proceedings of the Third DialDoc Workshop on Document-grounded Dialogue and Conversational Question Answering}, pages 80--85, Toronto, Canada. Association for Computational Linguistics.

\bibitem[{Sai et~al.(2019)Sai, Gupta, Khapra, and Srinivasan}]{sai2019reevaluating}
Ananya Sai, Mithun~Das Gupta, Mitesh~M. Khapra, and Mukundhan Srinivasan. 2019.
\newblock \href {https://arxiv.org/abs/1902.08832} {Re-evaluating {ADEM:} {A} deeper look at scoring dialogue responses}.
\newblock \emph{CoRR}, abs/1902.08832.

\bibitem[{Tack et~al.(2023)Tack, Kochmar, Yuan, Bibauw, and Piech}]{tack-etal-2023-bea}
Ana{\"\i}s Tack, Ekaterina Kochmar, Zheng Yuan, Serge Bibauw, and Chris Piech. 2023.
\newblock \href {https://doi.org/10.18653/v1/2023.bea-1.64} {The {BEA} 2023 shared task on generating {AI} teacher responses in educational dialogues}.
\newblock In \emph{Proceedings of the 18th Workshop on Innovative Use of NLP for Building Educational Applications (BEA 2023)}, pages 785--795, Toronto, Canada. Association for Computational Linguistics.

\bibitem[{Touvron et~al.(2023)Touvron, Martin, Stone, Albert, Almahairi, Babaei, Bashlykov, Batra, Bhargava, Bhosale, Bikel, Blecher, Ferrer, Chen, Cucurull, Esiobu, Fernandes, Fu, Fu, Fuller, Gao, Goswami, Goyal, Hartshorn, Hosseini, Hou, Inan, Kardas, Kerkez, Khabsa, Kloumann, Korenev, Koura, Lachaux, Lavril, Lee, Liskovich, Lu, Mao, Martinet, Mihaylov, Mishra, Molybog, Nie, Poulton, Reizenstein, Rungta, Saladi, Schelten, Silva, Smith, Subramanian, Tan, Tang, Taylor, Williams, Kuan, Xu, Yan, Zarov, Zhang, Fan, Kambadur, Narang, Rodriguez, Stojnic, Edunov, and Scialom}]{touvron2023llama2openfoundation}
Hugo Touvron, Louis Martin, Kevin Stone, Peter Albert, Amjad Almahairi, Yasmine Babaei, Nikolay Bashlykov, Soumya Batra, Prajjwal Bhargava, Shruti Bhosale, Dan Bikel, Lukas Blecher, Cristian~Canton Ferrer, Moya Chen, Guillem Cucurull, David Esiobu, Jude Fernandes, Jeremy Fu, Wenyin Fu, Brian Fuller, Cynthia Gao, Vedanuj Goswami, Naman Goyal, Anthony Hartshorn, Saghar Hosseini, Rui Hou, Hakan Inan, Marcin Kardas, Viktor Kerkez, Madian Khabsa, Isabel Kloumann, Artem Korenev, Punit~Singh Koura, Marie-Anne Lachaux, Thibaut Lavril, Jenya Lee, Diana Liskovich, Yinghai Lu, Yuning Mao, Xavier Martinet, Todor Mihaylov, Pushkar Mishra, Igor Molybog, Yixin Nie, Andrew Poulton, Jeremy Reizenstein, Rashi Rungta, Kalyan Saladi, Alan Schelten, Ruan Silva, Eric~Michael Smith, Ranjan Subramanian, Xiaoqing~Ellen Tan, Binh Tang, Ross Taylor, Adina Williams, Jian~Xiang Kuan, Puxin Xu, Zheng Yan, Iliyan Zarov, Yuchen Zhang, Angela Fan, Melanie Kambadur, Sharan Narang, Aurelien Rodriguez, Robert Stojnic, Sergey Edunov, and Thomas
  Scialom. 2023.
\newblock \href {https://arxiv.org/abs/2307.09288} {Llama 2: Open foundation and fine-tuned chat models}.
\newblock \emph{Preprint}, arXiv:2307.09288.

\bibitem[{Wang et~al.(2023)Wang, Liang, Meng, Sun, Shi, Li, Xu, Qu, and Zhou}]{wang-etal-2023-chatgpt}
Jiaan Wang, Yunlong Liang, Fandong Meng, Zengkui Sun, Haoxiang Shi, Zhixu Li, Jinan Xu, Jianfeng Qu, and Jie Zhou. 2023.
\newblock \href {https://aclanthology.org/2023.newsum-1.1} {Is {C}hat{GPT} a good {NLG} evaluator? a preliminary study}.
\newblock In \emph{Proceedings of the 4th New Frontiers in Summarization Workshop}, pages 1--11, Hybrid. Association for Computational Linguistics.

\bibitem[{Zhao et~al.(2020)Zhao, Lala, and Kawahara}]{zhao-etal-2020-designing}
Tianyu Zhao, Divesh Lala, and Tatsuya Kawahara. 2020.
\newblock \href {https://doi.org/10.18653/v1/2020.acl-main.4} {Designing precise and robust dialogue response evaluators}.
\newblock In \emph{Proceedings of the 58th Annual Meeting of the Association for Computational Linguistics}, pages 26--33, Online. Association for Computational Linguistics.

\bibitem[{Zhong et~al.(2022)Zhong, Liu, Yin, Mao, Jiao, Liu, Zhu, Ji, and Han}]{zhong-etal-2022-towards}
Ming Zhong, Yang Liu, Da~Yin, Yuning Mao, Yizhu Jiao, Pengfei Liu, Chenguang Zhu, Heng Ji, and Jiawei Han. 2022.
\newblock \href {https://doi.org/10.18653/v1/2022.emnlp-main.131} {Towards a unified multi-dimensional evaluator for text generation}.
\newblock In \emph{Proceedings of the 2022 Conference on Empirical Methods in Natural Language Processing}, pages 2023--2038, Abu Dhabi, United Arab Emirates. Association for Computational Linguistics.

\end{thebibliography}
\appendix
\onecolumn

\section{UniEval}

We made several changes to the original \uni to make it more suitable for the DailyDialog subset.  \uni was originally made for the Topical-Chat~\citep{mehri-eskenazi-2020-usr} benchmark, which is a grounded dialogue dataset with dialogue histories, candidate responses, and knowledge context in the form of a fact.  The \textbf{content} and \textbf{groundedness} submetrics both refer to this fact, so we experimented with altering the prompts to remove the reliance on this fact.  As \textbf{groundedness} judges if the candidate response correctly uses the fact, we judge this metric to be inextricable from the fact and remove it all together.  \textbf{Understandability} is a submetric that exists in the code, but not in the \uni paper, so we chose to exclude it as well.
The results of the ablation experiments are listed in Table \ref{tab:unieval}.

\label{sec:unieval}
\begin{table*}[!ht]
\begin{center}
\begin{tabular}{l|cccc}
      Metrics & Content & Naturalness & Relevance & Overall\\
    \hline
    \uni & 0.273 & 0.117 & 0.381 & 0.198  \\
     - Groundedness & 0.273  & 0.117 & 0.381 & 0.219 \\
     - Understandability & 0.273  & 0.117  & 0.381  & 0.236 \\
     + new content prompt  & 0.322  & 0.117  & 0.381  & 0.337 \\
    + weighted average  & 0.322  & 0.117 &  0.381  & 0.347 \\

\end{tabular}
\caption{Turn level Kendall's $\tau$ correlations with human judgment of different metrics on the DailyDialog subset for different versions of the \uni metric.}
\label{tab:unieval}

 \end{center}
\end{table*}

\onecolumn
\section{GPT Prompt}
\label{sec:gpt_appendix}
The full prompt for evaluation was as follows:

\begin{table*}[!h]
\small
\begin{center}
\begin{tabular}{p{\textwidth}}
You will be given a conversation between two individuals. You will then be given one potential response for the next turn in the conversation. \\
Your task is to rate the response on a series of metrics: content quality, grammaticality, and relevance.  Finally, you will assign an overall score (not an average). \\
Please make sure you read and understand these instructions carefully. Please keep this document open while reviewing, and refer to it as needed. \\
\\
Evaluation Criteria: \\
\\
Content Quality(1-5) - How compelling is the content of the response, and to what extent does it actively contribute to the ongoing conversation? \\
- A score of 1 (generic or boring) suggests that the response lacks interesting content and fails to contribute meaningfully to the conversation, potentially coming across as generic or dull.\\
- A score of 3 (moderately engaging) indicates that the response contains some interesting elements, contributing somewhat to the conversation, but there is room for improvement.\\
- A score of 5 (interesting and engaging) signifies that the content is exceptionally interesting, capturing attention and actively enhancing the overall conversation, demonstrating a high level of originality and contribution.\\
\\
Grammaticality(1-5) - How grammatical is the response? Consider only the response itself, not the conversation history.\\
- A score of 1 (ungrammatical) indicates that the response is confusing, lacks coherence, and is difficult to comprehend.\\
- A score of 3 (somewhat grammatical) suggests that the response is moderately clear but may contain some ambiguous or convoluted elements.\\
- A score of 5 (grammatical) indicates that the response is exceptionally clear, logically organized, and easy to understand.\\
\\
Relevance (1-5) - How well does the response align with the current conversational context and contribute meaningfully to the ongoing discourse? Pay close attention to the speaker.\\
- A score of 1 (irrelevant) suggests that the response is not related to the current conversation or significantly deviates from the established context.\\
- A score of 3 (somewhat relevant) indicates a partial alignment with the conversation but may contain elements that are not entirely pertinent to the ongoing discourse.\\
- A score of 5 (highly relevant) signifies that the response is directly related to the current conversation, seamlessly fitting into the established context without introducing unnecessary tangents or needless repetition.\\
\\
Overall Score (1-5) - How would you rate the response overall?.\\
- A score of 1 (poor) indicates that the response is of unrelated, boring, generic, or nonsensical.\\
- A score of 3 (average) suggests that the response is reasonably appropriate for the conversation, passably understandable, and somewhat interesting.\\
- A score of 5 (excellent) signifies that the response is exceptionally interesting and engaging, relevant to the conversation, and easy to understand.\\
\\
Conversation History:\\
(The following is a conversation between Alice and Bob.)\\
\textit{Alice: Well, how does it look?}\\
\textit{Bob: It's a perfect fit.}\\
\textit{Alice: Let me pay for it now.}\\
\\
Response:\\
\textit{Bob: Cash, credit card, or debit card?}\\
\\
Evaluation Form (scores ONLY):\\

\end{tabular}
\label{tab:gpt_prompt}

 \end{center}
\end{table*}
\newpage
\section{GPT Models}
\label{sec:gpt_results_app}
\begin{table*}[!ht]
\small
\begin{center}
\begin{tabular}{|l|rr|rr|r|}
    \hline
    & \multicolumn{2}{|c|}{\textbf{gpt-3.5-turbo}} & \multicolumn{2}{c|}{\textbf{gpt-4-preview}} & \textbf{gpt-4o}\\
    Attacks & 1106 & 0125 & 1106 & 0125 & 2024-05-13\\
    \hline
    \multicolumn{6}{|l|}{\textit{Speaker Tags} }\\
    \hline
    \textbf{teacher:} prefix & 0.38 & 0.04 & 0.17 & 0.27 & \textbf{0.01} \\
    \textbf{agent:} prefix & 0.55 & \textbf{0.09} & 0.67 & 0.79 & 0.42 \\
    \textbf{user:} prefix & 0.75 & \textbf{0.19} & 0.67 & 0.86 & 0.39\\
    \hline 
    \multicolumn{6}{|l|}{\textit{Static Responses}} \\
    \hline
    ``Hello'' & 0.04 & 0.03 & \textbf{0.01} & \textbf{0.01} & \textbf{0.01}\\ 
    ``I don't know'' & \textbf{0.01} & 0.02 & 0.02 & \textbf{0.01} & 0.02\\ 
    ``I don't know, what do you think?'' & 0.18 & 0.11 & 0.3 & \textbf{0.02} & 0.04\\
    ``I don't know, what do you think? I think'' & 0.01 & \textbf{0.00} & 0.01 & \textbf{0.00} & 0.01\\
    ``I'm sorry, can you repeat?'' & 0.19 & 0.10 & 0.03 & \textbf{0.02} & 0.03\\
    ``I will do'' & 0.07 & \textbf{0.00} & 0.01 & 0.02 & 0.01\\
    ``fantastic! how are you?'' & 0.03 & 0.05 & 0.03 & \textbf{0.02} & \textbf{0.02}\\
    \hline
    \multicolumn{6}{|l|}{\textit{Ungrammatical Responses}} \\
    \hline
    No punctuation & 0.48 & 0.21 & 0.32 & 0.36 & \textbf{0.19} \\
    No stopwords  & 0.16 & 0.07 & 0.09 & 0.09 & \textbf{0.00} \\
    Only nouns and verbs & 0.10 & 0.05 & 0.06 & 0.06 & \textbf{0.04} \\
    Only nouns & 0.23 & 0.06 & \textbf{0.04} & 0.05 & 0.07  \\
    Jumbled words & 0.02 & 0.02 & 0.02 & 0.03 & \textbf{0.01} \\
    Reversed words & 0.02 & \textbf{0.01} & \textbf{0.01} & \textbf{0.01} & \textbf{0.01} \\
    Repeated words & 0.09 & 0.04 & 0.04 & \textbf{0.03} & \textbf{0.03} \\
    \hline
    \multicolumn{6}{|l|}{\textit{Context Repetition Responses}} \\
    \hline
    Previous utterance & 0.60 & 0.41 & \textbf{0.01} & \textbf{0.01} & \textbf{0.01}\\
    Previous utterance prefix & 0.78 & 0.55 & \textbf{0.08} & 0.17 & 0.11\\
    \hline
\end{tabular}
\caption{The success rate of each attack against different evaluation metrics using \prompteval with different snapshots of GPT on DailyDialog.  The model is listed on the top line, with the specific snapshot used below.  The lower the number, the more resistant to attack.  The result of the best performing metric for each attack is bold.}
\label{tab:gpt_attacks}
 \end{center}
\end{table*}

\begin{table*}[!ht]
\begin{center}
\small
\begin{tabular}{l|cccc}
      \textbf{Metrics} & Content & Grammar & Relevance & Overall\\
    \hline
    gpt-3.5-turbo-1106  & 0.302  & 0.437  & 0.525  & 0.484\\
    gpt-3.5-turbo-0125  & 0.352  & 0.408  & 0.494 & 0.422\\
    \hline
    gpt-4-1106-preview  & 0.411  & 0.477  & 0.583  & 0.548 \\ 
    gpt-4-0125-preview  & 0.431 & 0.489  & 0.619  & 0.560 \\
    \hline
    gpt-4o-2024-05-13 & \textbf{0.437} &  \textbf{0.505}  & \textbf{0.627} & \textbf{0.562} \\

\end{tabular}
\caption{Turn level Kendall's $\tau$ correlations of different metrics on the DailyDialog subset for different snapshots of GPT. The highest value in each column is bolded.}
 \label{tab:gpt_version_correlation}

 \end{center}
\end{table*}

\end{document}